\pdfoutput=1
\documentclass[letterpaper]{article} 
\usepackage{aaai23}  
\usepackage{times}  
\usepackage{helvet}  
\usepackage{courier}  
\usepackage[hyphens]{url}  
\usepackage{graphicx} 
\usepackage{natbib}  
\usepackage{caption} 
\usepackage{algorithm}
\usepackage{algorithmic}

\usepackage{newfloat}
\usepackage{listings}
\DeclareCaptionStyle{ruled}{labelfont=normalfont,labelsep=colon,strut=off} 
\floatstyle{ruled}
\newfloat{listing}{tb}{lst}{}
\floatname{listing}{Listing}
\usepackage{amsmath}
\usepackage{amsfonts}

\title{StyleFaceV: Face Video Generation \\
via Decomposing and Recomposing Pretrained StyleGAN3}
\author{
       Haonan Qiu,\textsuperscript{\rm 1}
       Yuming Jiang,\textsuperscript{\rm 1}
       Hang Zhou,\textsuperscript{\rm 2}\\
       Wayne Wu,\textsuperscript{\rm 3}
       Ziwei Liu \textsuperscript{\rm 1}
   }
   \affiliations {
       \textsuperscript{\rm 1}S-Lab, Nanyang Technological University,\\
       \textsuperscript{\rm 2}The Chinese University of Hong Kong,\\
       \textsuperscript{\rm 3}SenseTime Research\\
       \{haonan002, yuming002, ziwei.liu\}@ntu.edu.sg, zhouhang@link.cuhk.edu.hk, wuwenyan0503@gmail.com
}

\usepackage{bibentry}

\begin{document}

\twocolumn[{%
    \renewcommand\twocolumn[1][]{#1}%
    \maketitle
    \begin{center}
        \centering
        \includegraphics[width=0.99\textwidth]{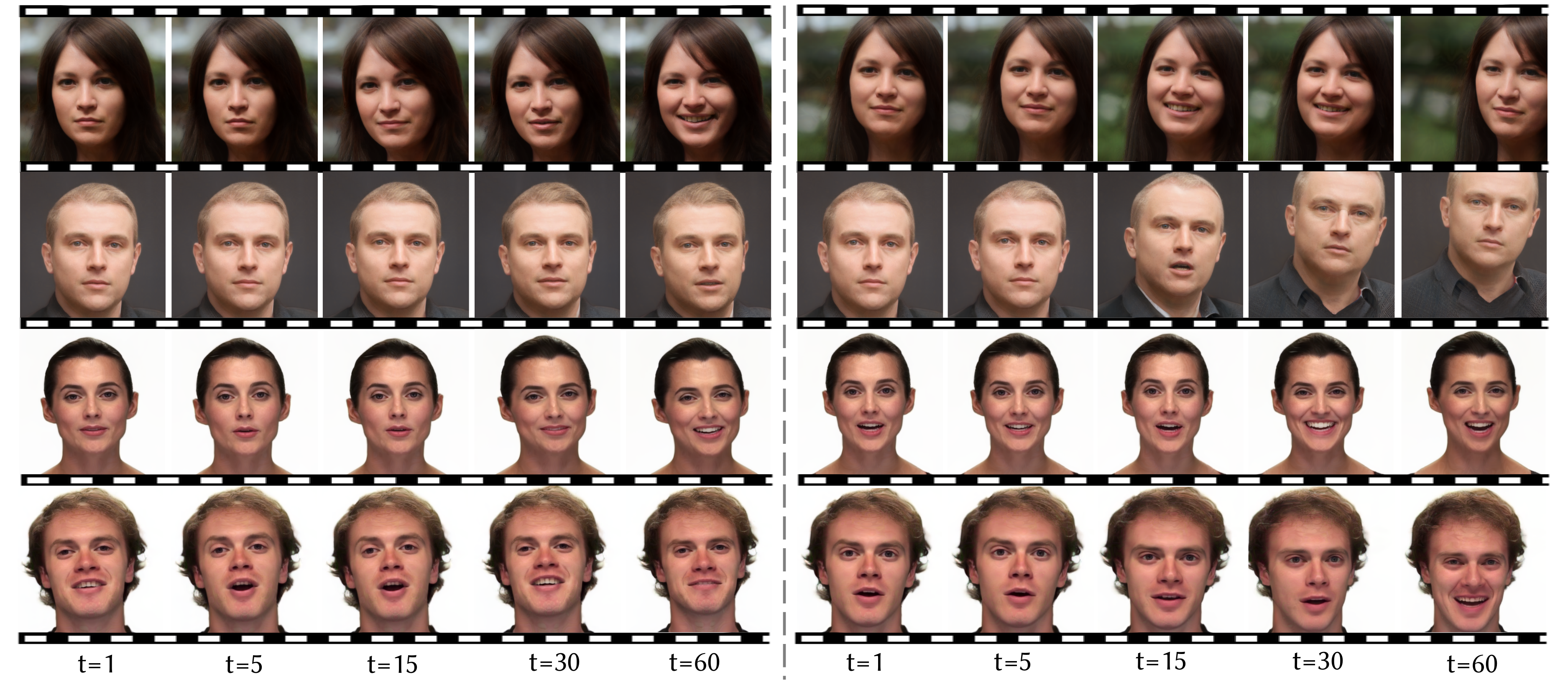}
        \vspace{-10pt}
        \captionof{figure}{\textbf{Face Video Generation with Realistic and Diverse Appearances and Motions.} Our proposed \textbf{StyleFaceV} is able to generate videos by sampling appearance representations and pose representations.
        For each face identity, we show two video clips. The left one is with small motions and the right one is with large motions.
        The face identity is kept even in large motions.}
        \label{fig:teaser}
    \end{center}%
        }]

\begin{abstract}
  Realistic generative face video synthesis has long been a pursuit in both computer vision and graphics community.
  However, existing face video generation methods tend to produce low-quality frames with drifted facial identities and unnatural movements.
  To tackle these challenges, we propose a principled framework named \textbf{StyleFaceV}, which produces high-fidelity identity-preserving face videos with vivid movements. Our core insight is to \emph{decompose appearance and pose information and recompose them in the latent space of StyleGAN3 to produce stable and dynamic results.}
  Specifically, StyleGAN3 provides strong priors for high-fidelity facial image generation, but the latent space is intrinsically entangled. By carefully examining its latent properties, we propose our decomposition and recomposition designs which allow for the disentangled combination of facial appearance and movements. 
  Moreover, a temporal-dependent model is built upon the decomposed latent features, and samples reasonable sequences of motions that are capable of generating realistic and temporally coherent face videos.
  Particularly, our pipeline is trained with a joint training strategy on both static images and high-quality video data, which is of higher data efficiency.
  Extensive experiments demonstrate that our framework achieves state-of-the-art face video generation results both qualitatively and quantitatively.
  Notably, StyleFaceV is capable of generating realistic $1024 \times 1024$ face videos even without high-resolution training videos.
\end{abstract}
\section{Introduction}

Face generation has been a long-standing research topic in vision and graphics. As humans perceive the world from a video-point of view, the exploration of video face generation is of great importance.
Despite the great progress~\cite{goodfellow2014generative,karras2019style,karras2020analyzing,karras2020training} in generating static face images, the task of face video generation is less explored due to the extra complexity introduced by the temporal dimension. 
Compared to a face image, we expect a generated face video to possess three desired properties: \textbf{1)} \emph{High fidelity.} The face of each frame should be photo-realistic. \textbf{2)} \emph{Identity preservation.} The same face identity should be kept among frames in the generated video. \textbf{3)} \emph{Vivid dynamics.} A natural face video is supposed to contain a reasonable sequence of motions.

However, current face video generation methods still struggle to generate visually satisfying videos that meet all desired properties above. 
Most previous works design specific end-to-end generative models~\cite{tian2021good,tulyakov2018mocogan,clark2019adversarial,skorokhodov2021stylegan,yu2022generating} that target directly at video generation. 
Such designs are expected to achieve both spatial quality and temporal coherence in a single framework trained from scratch, imposing great burden to the network.
In another fashion, some recent works~\cite{fox2021stylevideogan,tian2021good} synthesize videos by employing a pre-trained StyleGAN and finding the corresponding sequence of latent codes. But those methods generate videos by manipulating the entangled latent codes thus making identity easily change with head movements.

To address these challenges, we propose a new framework, named \textbf{StyleFaceV} with three desired properties: photo-realism, identity preservation and vivid dynamics. 
In Fig.~\ref{fig:teaser}, we show some videos generated by our framework. The two videos in each row share the same face identity but with different facial movements.
Our framework uses the StyleGAN3~\cite{karras2021alias} to synthesize a sequence of face frames to obtain the final generated video. With the powerful image generator, the key to generate a realistic video is: \textit{How can we generate a temporally coherent video with the same identity by traversing through the latent space of StyleGAN3?}
Our core insight is to decompose the latent space of the StyleGAN3 into the appearance and pose information via \textit{a decomposition and recomposition pipeline}.

The decomposition module extracts pose and appearance information from the images synthesized by StyleGAN3. 
We design two extraction networks to extract pose features and appearance features, respectively.
With the decomposed features, the recomposition module is employed to fuse them back to the latent codes as the inputs to StyleGAN3. The fused latent codes are then used to generate the facial images.
With the decomposition and recomposition pipeline, we can easily sample a well disentangled sequence of latent codes with the same identity by sampling the pose information while keeping the appearance information fixed.
The sampling of pose sequences is achieved by an additional LSTM~\cite{hochreiter1997long} module. 

To train the decomposition and recomposition pipeline, we need faces with the same identity but different poses. Such paired data can be easily obtained from face videos. However, large-scale and high-quality video datasets are lacking. To alleviate the problems caused by the lack of high-quality videos, we propose the joint training strategy, which utilizes the rich image priors captured in the well pretrained StyleGAN3 model as well as the limited video data. 
Specifically, we do the self augmentation by synthesizing a new frame with a randomly translated and rotated face through the affine layer embedded in StyleGAN3 to simulate paired frames of face videos.
The use of synthetic data eases the requirements on the video dataset. 
Notably, this joint training design enables our model to generate realistic $1024 \times 1024$ face videos even without accessing high-resolution training videos.

In summary, our main contributions include:
\begin{itemize}
  \item We propose a novel framework \textbf{StyleFaceV} which decomposes and recomposes pretrained StyleGAN3 to generate new face videos via sampling decomposed pose sequences.
  \item We design a joint training strategy which utilizes information from both the image domain and video domain.
  \item Our work exhibits that latent space of StyleGAN3 contains a sequence of latent codes which is able to generate a temporally coherent face video with vivid dynamics.
\end{itemize}
\begin{figure*}[t]
\centering
\includegraphics[width=1.0\linewidth]{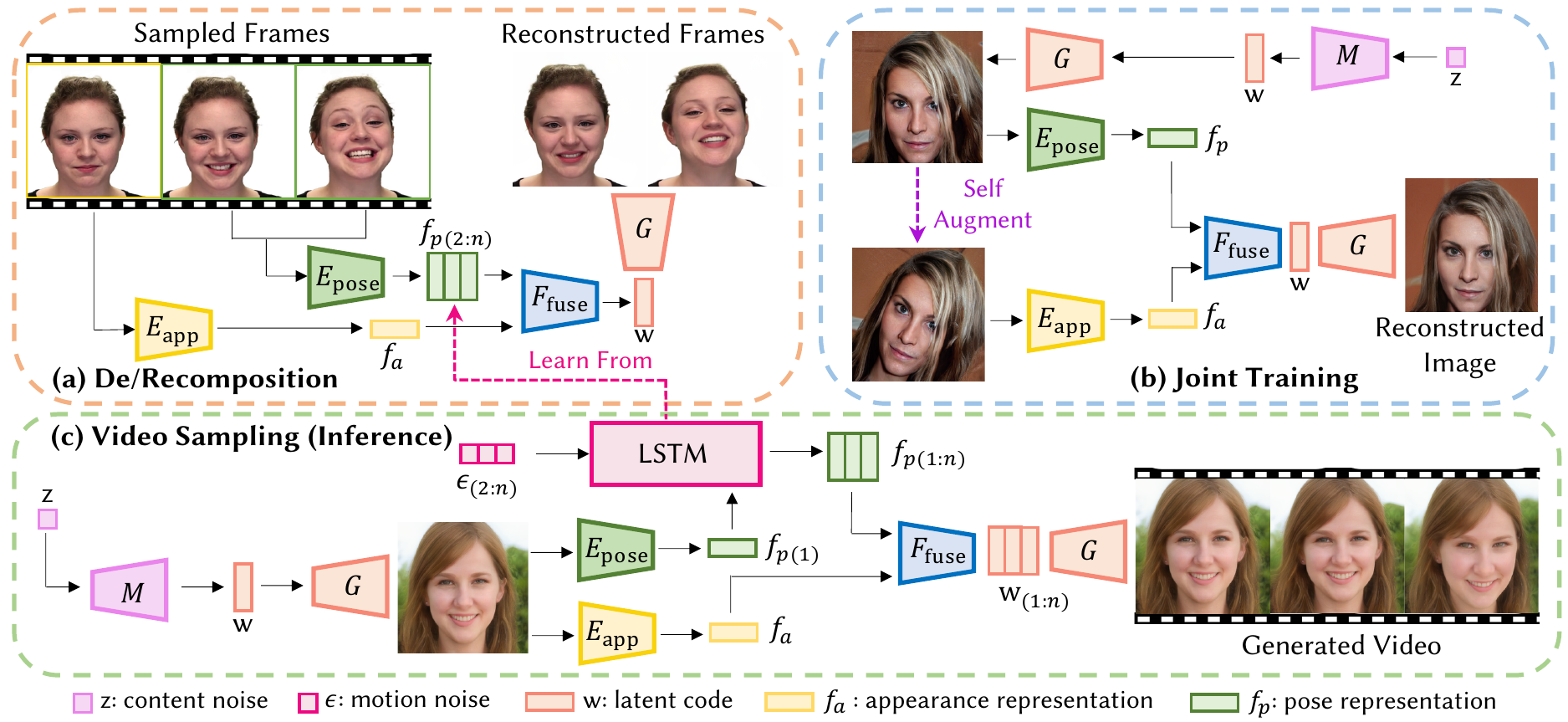}
\vspace{-15pt}
\caption{\textbf{Overview of Our Proposed StyleFaceV.} (a) Our proposed StyleFaceV decomposes the image into an appearance representation $f_{a}$ and a pose representation $f_{p}$. Then we re-compose both representations to get the intermediate embedding $\mathbf{w}$ as the input of the synthesis network $G$. 
(b) Besides video training data, joint training strategy also samples generated images for training to utilize rich image priors captured in the well pretrained StyleGAN3 model by sampling initial content noise $\mathbf{z}$. 
To simulate paired frames with different poses but the same appearance, self-augmentation is used to randomly transform the generated image through the embedded function of StyleGAN3.
(c) To generate videos from scratch, an LSTM-based motion sampling module is trained to sample a sequence of pose representations $\{f_{p(1)},...,f_{p(n)}\}$ from motion noise $\mathbf{\epsilon}$, which is then re-composed with a randomly sampled appearance representation $f_{a}$ to generate a sequence of frames.} 
\vspace{-15pt}
\label{fig:pipeline}
\end{figure*}

\section{Related Work}

\noindent\textbf{StyleGAN Models.}
With the success of Generative Adversarial Networks (GAN) ~\cite{goodfellow2014generative}, generative models have become a popular technique to generate natural images. After that, various GAN architectures \cite{karras2017progressive, miyato2018spectral, brock2018large} are proposed to stabilize the training process and the quality of generated images is greatly improved. After StyleGAN series \cite{karras2019style,karras2020analyzing,karras2020training} have achieved high-resolution generation and style controlling, \cite{karras2021alias} propose StyleGAN3 to solve the ``texture sticking'' problem, paving the way for generative models better suited for video and animation.
Besides generating images from scratch, those pre-trained image generators are also useful for performing several image manipulation and restoration tasks by utilizing captured natural image priors, such as image colorization \cite{pan2021exploiting,wu2021colorization}, super-resolution \cite{chan2021glean,wang2021gfpgan,menon2020pulse} and facial image editing \cite{shen2020interfacegan,patashnik2021styleclip,jiang2021talk}.
In our proposed StyleFaceV framework, we also utilized a pre-trained StyleGAN3 to render photo-realistic face videos.

\noindent\textbf{Video Generation.}
Different from image generation which only requires sampling at the spatial level, video generation involves an additional sampling at temporal level. Many works of video synthesis have achieved impressive results, \emph{e.g.}, video-to-video translation with given segmentation masks ~\cite{wang2018video,wang2019few} or human poses ~\cite{chan2019everybody}, and face reenactment with given both human identities and driven motions ~\cite{siarohin2019first, zhou2021pose}.
However, in the unconditional setting, generating videos from scratch remains still unresolved.
Early works \cite{tulyakov2018mocogan,saito2017temporal,hyun2021self,aich2020non,munoz2021temporal,saito2020train} proposed to use a single framework to synthesize low-resolution videos by sampling from some noise. \cite{tulyakov2018mocogan} decompose noises in the content domain and motion domain. The content noise controls the content of the synthesized video, while motion noises handle the temporal consistency.
Recently, MoCoGAN-HD \cite{tian2021good}, and StyleVideoGAN \cite{fox2021stylevideogan} scaled up the resolution of synthesized videos by employing pre-trained StyleGAN. 
MoCoGAN-HD and StyleVideoGAN proposed to sample a sequence of latent codes, which were then fed into pre-trained StyleGAN to synthesize a sequence of images. The pose information and identity information were mixed in one single latent code. 
Different from these works, our methods explicitly decompose the input to the generator into two branches, \emph{i.e.}, pose and appearance, via a decomposition loss.
DI-GAN \cite{yu2022generating} introduces an INR-based video generator that improves motion dynamics by manipulating space and time coordinates separately. However, it has a high computational cost and suffers sticking phenomenon. 
StyleGAN-V \cite{skorokhodov2021stylegan} modified architectures of the generator and discriminator of StyleGAN by introducing a pose branch to inputs. 
Both DI-GAN and StyleGAN-V require the retraining of generators on video datasets, making the faces in synthesized videos restricted to the domain of the training set. By contrast, our method utilizes image priors captured in the pre-trained StyleGAN3 and synthesizes facial videos with diverse identities.

\section{Our Approach}

Our framework is built on top of powerful pre-trained image generator StyleGAN3~\cite{karras2021alias}, which provides high quality face frames. As shown in Fig.~\ref{fig:pipeline}, a decomposition and recomposition pipeline is used to find some sequences of latent codes from the latent space of StyleGAN3 to generate a temporally coherent sequence of face images with the same identity. Finally, the motion sampler models the distribution of natural movements and samples a sequence of poses to drive the movements of faces.

\subsection{Pre-Trained Image Generator}

The recently proposed StyleGAN3~\cite{karras2021alias} does not suffer from the ``texture sticking'' problem and thus paves the way for generating videos by pre-trained image generators. Among several variants of StyleGAN3, only StyleGAN3-R is equivalent to the rotation transformations. Considering face rotations are commonly observed in talking-face videos, we adopt the pre-trained StyleGAN3-R as our image generator. It contains a mapping network $M$ that maps the content noise $\mathbf{z}$ into an intermediate latent space $\mathcal{W}$, and a synthesis network $G$ that synthesizes image $\mathbf{x} \in \mathcal{I}$ from the intermediate embedding $\mathbf{w} \in \mathcal{W}$:
\begin{equation}
\begin{aligned}
\mathbf{w}
= M(\mathbf{z}),
\mathbf{x}
= G(\mathbf{w}).
\end{aligned}
\label{eqn:method-stylegan}
\end{equation}
We slightly fine-tune the generator on video dataset so that it can generate images from the distribution of video dataset.

In the unconditional setting, face video generation task aims to generate a realistic face video sequence $\mathbf{v} \in \mathcal{V}$ from scratch. Regarding the generated image as a single frame, a sequence of generated images can form the final synthesized video $\mathbf{v} =\{\mathbf{x}_1, \mathbf{x}_2, ...,\mathbf{x}_\text{n}\}$.
Therefore, the rest problem is how to find a sequence of latent codes from the latent space of StyleGAN3 to generate a temporally coherent sequence of face images with the same identity.

\subsection{Decomposition and Recomposition}

To find the path in the latent space of StyleGAN3 which is able to change the face actions and movements without changing the identity, we propose a decomposition and recomposition pipeline as shown in Fig.~\ref{fig:pipeline}(a).
In this pipeline, we decompose a face image into an appearance representation $f_{a} = E_\text{app}(\mathbf{x})$ and a pose representation $f_{p} = E_\text{pose}(\mathbf{x})$. Then we can simply re-compose $f_{a}$ and $f_{p}$ to get the intermediate embedding $\mathbf{w}$ as the input of the synthesis network $G$ by the re-composition network $F_\text{fuse}$: 
\begin{equation}
\mathbf{w}
= F_\text{fuse}(f_{a}, f_{p}).
\label{eqn:method-fuse}
\end{equation}

This pipeline allows us to drive the face by changing the pose representation $f_{p}$ while the identity is kept by freezing the appearance representation $f_{a}$.

\begin{figure}[t]
\centering
\includegraphics[width=0.9\linewidth]{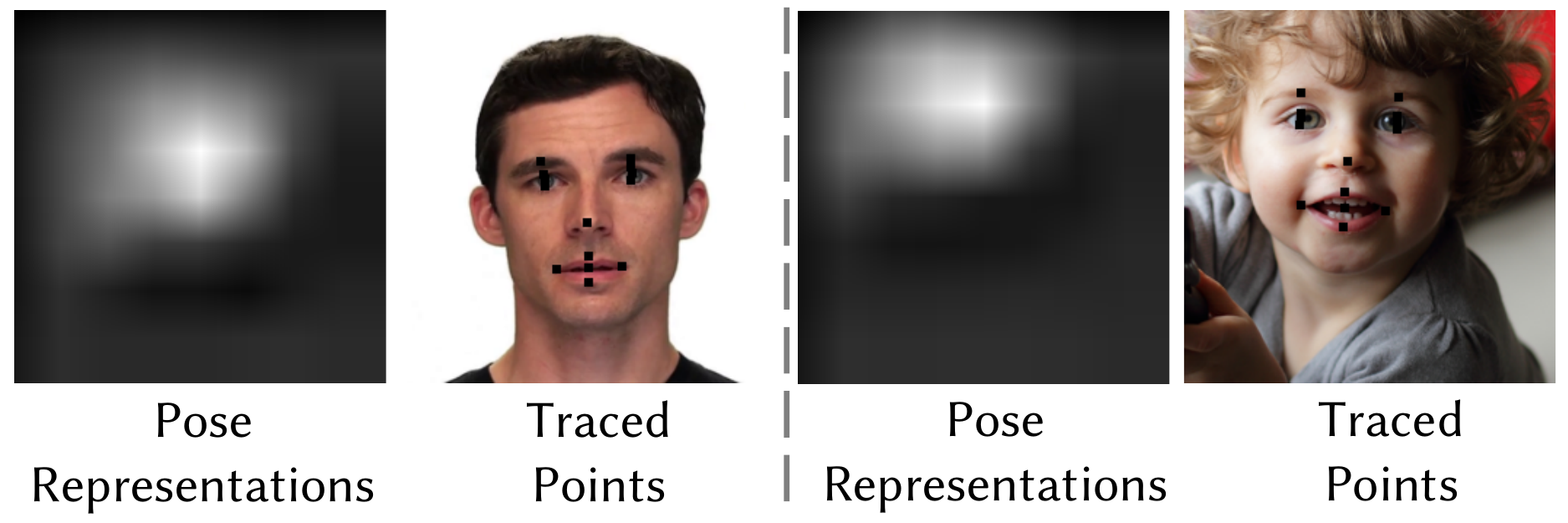}
\vspace{-10pt}
\caption{\textbf{Pose Extraction.} We visualize the pose representations for two persons and draw the traced points $E_\text{kp}(f_{p})$ on its corresponding image. 
}
\vspace{-15pt}
\label{fig:pose}
\end{figure}

\noindent\textbf{Pose Extraction.}
Pose representations in face videos mainly consist of two levels of information: \textbf{1)} overall head movement, \textbf{2)} local facial movement. Both of these two kinds of movements will be reflected in landmarks. Therefore, we use a pre-trained landmark detection model~\cite{Wang_2019_ICCV} to supervise the pose estimation. However, all predicted landmarks also contain the identity information, like face shape. To purify the pose information without reserving the identity information, our pose extractor $E_\text{pose}$ are only trained to reserve information of selected key face points $p^\text{target}$:
\begin{equation}
\begin{aligned}
f_{p} = E_\text{pose}(\mathbf{x}), 
\label{eqn:method-pose}
\end{aligned}
\end{equation}
\begin{equation}
\begin{aligned}
\mathcal{L}_\text{pose}
= \|E_\text{kp}(f_{p}) - p^\text{target}\|^2_2,
\label{eqn:method-pose-loss}
\end{aligned}
\end{equation}
where $E_\text{kp}$ is key face points predictor and $p^\text{target}$ presents targeted key face points selected from predicted landmarks from the pre-trained landmark detection model.
As shown in Fig.~\ref{fig:pose}, points $p^\text{target}$ only trace the position of facial attributes and their status (e.g., open/close for mouth). The number of key face points is significantly less than that of landmarks. In addition, more detailed pose information which is not obtained by landmarks will be supervised during the reconstruction stage.

\noindent\textbf{Appearance Extraction.}
Different from the pose representations, it is hard to directly define appearance representation. In this face video generation task, we assume appearance is not changed. Based on this assumption, if we sample two frames $\mathbf{x}_i$ and $\mathbf{x}_j$ from the same video, we can get:
\begin{equation}
\begin{aligned}
\mathbf{x}^{*}_j 
&= G(F_\text{fuse}(f_{a(j)}, f_{p(j)})\\
&= G(F_\text{fuse}(E_\text{app}(\mathbf{x}_i), E_\text{pose}(\mathbf{x}_j)).
\end{aligned}
\label{eqn:method-appearance}
\end{equation}

In theory, $E_\text{app}$ can be learned by making $\mathbf{x}^{*}_j$ close to $\mathbf{x}_j$. However, we find that networks can hardly converge if we train the framework together from scratch. It is mainly because $E_\text{app}$ and $E_\text{pose}$ disentangle with each other, and can not provide effective appearance/pose information if we randomly initialize the network. Therefore, we first train $E_\text{pose}$ with the single loss in Eq.~\ref{eqn:method-pose-loss}. After $E_\text{pose}$ is pretrained, we train the whole framework together.

\noindent\textbf{Objective Functions.}
To use the video dataset, each time we sample $n$ frames 
$\mathbf{x} = \{\mathbf{x}_{i}, ..., \mathbf{x}_{i+k\times(n-1)}\}$
with interval $k$ and predict each frame $\mathbf{x}$ through Eq.~\ref{eqn:method-appearance}. Here we simply set $k = 3$ to make sampled frames have obvious pose differences. By training to reconstruct frames with the same appearance, $E_\text{app}$ is guided to ignore the pose information while $E_\text{pose}$ learns to extract the differences among frames, which are pose representations in this work.
The overall function for training with video dataset is:
\begin{equation}
\mathcal{L}_\text{overall}
= {\lambda_\text{1}}\mathcal{L}_\text{1} + {\lambda_\text{per}}\mathcal{L}_\text{per} + {\lambda_\text{gan}}\mathcal{L}_\text{gan} + {\lambda_\text{pose}}\mathcal{L}_\text{pose},
\label{eqn:method-video-loss}
\end{equation}
where $\mathcal{L}_\text{1} = \|\mathbf{x^{*}} - \mathbf{x}\|^1_1$ and $\mathcal{L}_\text{per} = \|\text{VGG16}(\mathbf{x^{*}}) - \text{VGG16}(\mathbf{x})\|^1_1$ are constraints to guarantee the reconstruction quality in both pixel domain and perceptual feature domain extracted by VGG16~\cite{simonyan2014very}. $\mathcal{L}_\text{gan}$ is an adversarial loss~\cite{goodfellow2014generative} to enhance local details. $\mathcal{L}_\text{pose}$ is a pose training loss introduced in Eq.~\ref{eqn:method-pose-loss}. $\lambda$ is the weight for different losses when using video data. When the pipeline is able to do roughly good images reconstruction with decomposition, $\lambda$s for training with sampled images will decrease to make pose representations bring a finer control, which will be further analyzed in ablation study.

\noindent\textbf{Joint Training Strategy.}
A well-trained StyleGAN3 model contains rich image priors captured in the large-scale and high-quality face image datasets. Utilizing this feature, we can also sample identities from distribution of original image datasets. However, there is no shared appearance among sampled images and each image is the individual frame. To prevent the networks from being lazy to do only reconstruction without decomposing, we design the joint training strategy as shown in Fig.~\ref{fig:pipeline}(b). We firstly do the self augmentation through the affine layer embedded in StyleGAN3-R to produce a new frame with a randomly translated and rotated face while its face appearance is kept and background is still natural.
Then we use the original version to provide the pose information and the transformed variants to provide the appearance information. 

Specifically, instead of sampling two frames $\mathbf{x}_i$ and $\mathbf{x}_j$ from the same video in Eq.~\ref{eqn:method-appearance}, we use the equation below to get $\mathbf{x}_i$ and $\mathbf{x}_j$ from the StyleGAN3-R:
\begin{equation}
\mathbf{x}_i=G\left(\mathbf{w} ;\left(r, t_{x}, t_{y}\right)\right), \mathbf{x}_j=G\left(\mathbf{w} ;\left(0, 0, 0\right)\right),
\end{equation}
where $r$ is the random rotation angle, and $t_{x}$ , $t_{y}$ are the random translation parameters.

In addition, we add a self-supervised embedding loss function $\mathcal{L}_\text{w}$ to help the convergence of training:
\begin{equation}
\mathcal{L}_\text{w}
= \|\mathbf{w}^* - \mathbf{w}\|^1_1,
\label{eqn:method-w-loss}
\end{equation}
where $\mathbf{w} = M(\mathbf{z}),$ and $\mathbf{w}^* = F_\text{fuse}(E_\text{app}(\mathbf{x_i}), E_\text{pose}(\mathbf{x_j}))$. Other loss functions are the same as Eq.~\ref{eqn:method-video-loss}.

\begin{figure*}[t]
\centering
\includegraphics[width=0.98\linewidth]{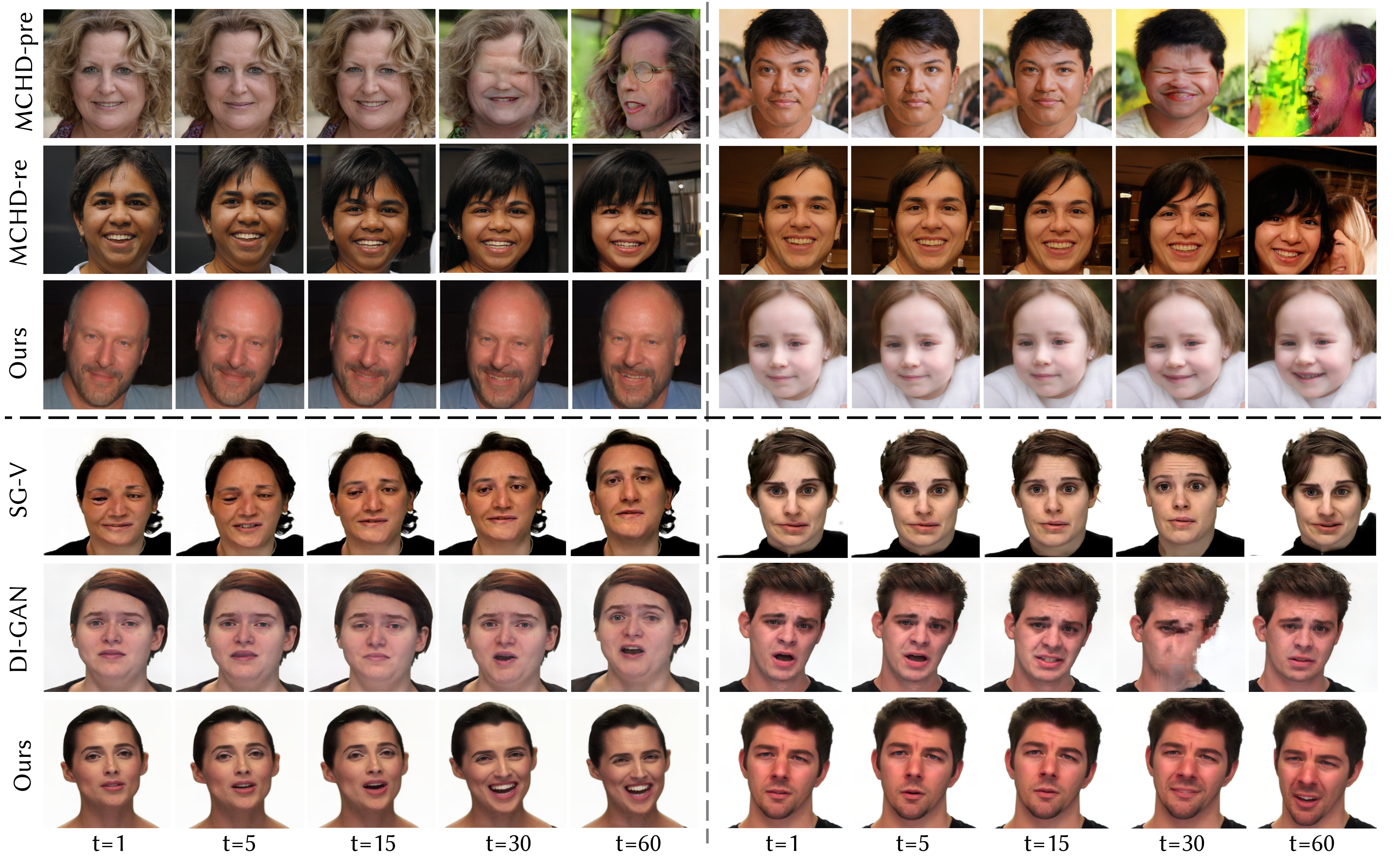}
\vspace{-10pt}
\caption{\textbf{Qualitative Comparison.} We compare our proposed StyleFaceV with some representative methods on the video generation task. Results of MCHD-pre and MCHD-re are from FFHQ domain, while results of SG-V and DI-GAN are from RAVDESS domain. Our StyleFaceV can generate videos by sampling from both domains and we compare them respectively. Qualitative comparison shows that our StyleFaceV outperforms all other methods in both quality and identity persevering.}
\vspace{-15pt}
\label{fig:comparison}
\end{figure*}

\subsection{Motion Sampling}

In order to generate videos from random motion noise $\mathbf{\epsilon}$, we design a motion sampling module to sample meaningful sequences of pose representations. Here we choose LSTM~\cite{hochreiter1997long} to generate later poses of $n-1$ frames $\{f_{p(2)},...,f_{p(n)}\}$ by giving the initial pose $f_{p(1)}$ and sampled motion noises $\{\mathbf{\epsilon}_{2},...,\mathbf{\epsilon}_{n}\}$: 
\begin{equation}
\{f_{p(2)},...,f_{p(n)}\} = \text{LSTM}(f_{p(1)}, \{\mathbf{\epsilon}_{2},...,\mathbf{\epsilon}_{n}\}).
\label{eqn:method-lstm}
\end{equation}

Because we have no ground truth for generated results, we use a 2D adversarial loss $\mathcal{L}_\text{gan}^\text{2D}$ to make the single generated pose representation fit the distribution of real videos and a 3D adversarial loss $\mathcal{L}_\text{gan}^\text{3D}$ whose inputs are consecutive pose representations to constrain the temporal consistency:
\begin{equation}
\begin{aligned}
\mathcal{L}_\text{gan}^\text{3D}
&=\mathbb{E}\left[\log D_{\mathrm{v}}(\{\mathbf{x}_1,\mathbf{x}_2,...,\mathbf{x}_\text{n}\})\right]\\
&+\mathbb{E}\left[\log \left(1-D_{\mathrm{v}}\left(G(F_\text{fuse}(f_{a(1)}, \{f_{p(1)},...,f_{p(n)}\}))\right)\right)\right],
\label{eqn:method-gan3d}
\end{aligned}
\end{equation}
where $D_{\mathrm{v}}$ is the video discriminator to predict whether a video is real or fake according to the given sequence.

Additionally, we add a diversity loss $\mathcal{L}_{\text{d}}$ to make the generated pose representations $p$ vary with the given motion noises $\epsilon$:
\begin{equation}
\mathcal{L}_{\text{d}}=-\frac{1}{n-1} \sum_{t=2}^{n}
(\text{FC}(\mathbf{h}_{t})^{T} \mathbf{\epsilon}_{t} /\|(\text{FC}(\mathbf{h}_{t}))\|\|\mathbf{\epsilon}_{t}\|),
\label{eqn:method-diversity}
\end{equation}
where $\text{FC}(\cdot)$ is a fully connected layer and $\mathbf{h}$ is the hidden feature. 

For the motion sampler of $256 \times 256$ resolution, we generate final videos with sampled pose sequence and apply supervisions on image/video representations because we find this operation brings more diverse motions.
But for $1024 \times 1024$ resolution, all losses are added on pose representations only, which are much smaller than image space, saving a large amount of computation cost. This design allows the training process to supervise a longer sequence.
 
\section{Experiments}

\subsection{Joint Unaligned Dataset}

Previous face video generation methods~\cite{tulyakov2018mocogan, saito2020train, skorokhodov2021stylegan} usually use FaceForensics~\cite{roessler2019faceforensicspp} as the video dataset and are able to generate good videos whose identities are from FaceForensics. However, StyleGAN-V~\cite{skorokhodov2021stylegan} points out the limitation that only $700$ identities of FaceForensics can be sampled. To overcome this problem, we propose the joint training strategy. We choose unaligned FFHQ~\cite{karras2019style} as image dataset that provides tens of thousands of identities for sampling and RAVDESS~\cite{livingstone2018ryerson} as video dataset which provides $2451$ high-quality vivid face motion sequences. 
Compared to FaceForensics, RAVDESS has more various motions and better video quality thus having a smaller gap with the image dataset FFHQ, helping the convergence of joint training.

In addition, many previous video generation methods~\cite{tulyakov2018mocogan, saito2020train, skorokhodov2021stylegan} stick the face in the center by aligning the face bounding box when preprocessing face video data. It erases natural head movement thus reducing the difficulty of face generation. However, this operation brings obvious face distortions and shaking among frames. In this paper, we use the setting~\cite{siarohin2019first} and allow the face to freely moved in a fixed window. It is more consistent with real-world face videos because, in real-world face videos, the camera is usually fixed.

\noindent\textbf{Evaluation Metrics.}
For evaluation, we report Frechet Inception Distance (FID)~\cite{heusel2017gans} and Frechet Video Distance (FVD)~\cite{unterthiner2018towards}. In this paper, we use two versions of FID: FID-RAVDESS and FID-Mixture, where FID-RAVDESS is computed against RAVDESS dataset and FID-Mixture is calculated against a mixture dataset (RAVDESS and FFHQ). We also conduct a user study to evaluate the quality of generated videos. We mix our generated videos with those generated by baselines because paired comparison is not supported in unconditional generation. A total of $30$ users were asked to give two scores ($1$-$5$, the best score is $5$) for the naturalism of movements and the identity preservation throughout the movements. 

\begin{table}[t]
\begin{center}
\caption{\textbf{Quantitative Comparisons on FID and FVD Score.} We compute the FID-RAVDESS between the generated frames and real frames from RAVDESS. FID-Mixture is calculated against a mixture dataset. FVD is computed between synthesized videos and videos in RAVDESS dataset.}
\vspace{-10pt}
\label{table:fid_fvd}
\scalebox{0.85}{\begin{tabular}{l|c|c|c}
\hline\noalign{\smallskip}
Methods & FID-RAVDESS ($\downarrow$) & FID-Mixture ($\downarrow$) & FVD ($\downarrow$) \\
\noalign{\smallskip}
\hline
\noalign{\smallskip}
MCHD-pre  & 174.44 & 53.53 & 1425.77 \\
MCHD-re & 146.08 & 73.73 & 1350.67 \\
StyleGAN-V & 17.93 & 95.35 & 171.60s \\
DIGAN & \textbf{12.02} & 77.01 & 142.08 \\
StyleFaceV & 15.42 & \textbf{25.31} & \textbf{118.72}
\\
\hline
\end{tabular}}
\end{center}
\vspace{-15pt}
\end{table}
\setlength{\tabcolsep}{1.4pt}

\subsection{Experimental Settings}

We compare our proposed StyleFaceV with the following representative methods on the face video generation task. Compared to previous sticking setting, unaligned face generation is more challenging. Considering huge computational costs for DI-GAN, all results are fixed with $60$ frames for the fair comparison.

\noindent\textbf{MoCoGAN-HD}~\cite{tian2021good} uses a pre-trained image generator and synthesizes videos by modifying the embedding in latent space.
We compare with two versions of MoCoGAN-HD: \textit{1) MCHD-pre} uses the released pretrained model which supports sampling identites from FFHQ domain. \textit{2) MCHD-re} retrains the models using StyleGAN3-R~\cite{karras2021alias} with video dataset RAVDESS~\cite{livingstone2018ryerson} for a fair comparison, as the original version is built upon StyleGAN2~\cite{karras2020analyzing} which does not support various unaligned face poses.

\noindent\textbf{DI-GAN}~\cite{yu2022generating} uses an INR-based video generator to improve the motion dynamics by manipulating the space and time coordinates separately. Because it only supports the video dataset, we retrain it on RAVDESS dataset. The number of training frames is set to $60$ which is the same as the testing phase to avoid the collapse.

\noindent\textbf{StyleGAN-V (SG-V)}~\cite{skorokhodov2021stylegan} builds the model on top of StyleGAN2, and redesigns its generator and discriminator networks for video synthesis. We also retrain it on RAVDESS dataset.

\subsection{Qualitative Comparison}

As presented in Fig.~\ref{fig:comparison}, MCHD-pre always generates faces located at the center of the image because it is based on StyleGAN2. In addition, generated results will crash with the time going as this model is pre-trained on $16$ frames setting. Our retrained MCHD-re improves those problems by using StyleGAN3. Meanwhile, video dataset RAVDESS makes MCHD-re learn the face motion. However, the man in the second video clip of MCHD-re gradually becomes a woman with the pose movement. The reason is that MCHD-re still moves along the entangled latent codes thus making identity change with other actions. Our method well decomposes the appearance representations and pose representations, preserving the identity throughout the pose movement. 

Another state-of-the-art method, DI-GAN exactly produces videos without identity changes. However, some of frames will be blurred or even crash. The most recent work StyleGAN-V does not suffer from the crash problem. However, its generated faces are easy to deform and distort. The poor stability is mainly because both DI-GAN and StyleGAN-V use a single end-to-end trained framework to handle both the quality and coherence of all synthesized frames, which gives a huge burden to the network.
Therefore, DI-GAN and StyleGAN-V cannot model the unaligned faces with movements.
In contrast, our framework uses a pre-trained StyleGAN3-R which only focuses on the quality of every single frame and uses other modules to undertake the duty of coherence. Results shown in Fig.~\ref{fig:comparison} exhibit that our StyleFaceV outperforms all other methods in both quality and identity keeping.

\begin{figure}[t]
\centering
\includegraphics[width=0.9\linewidth]{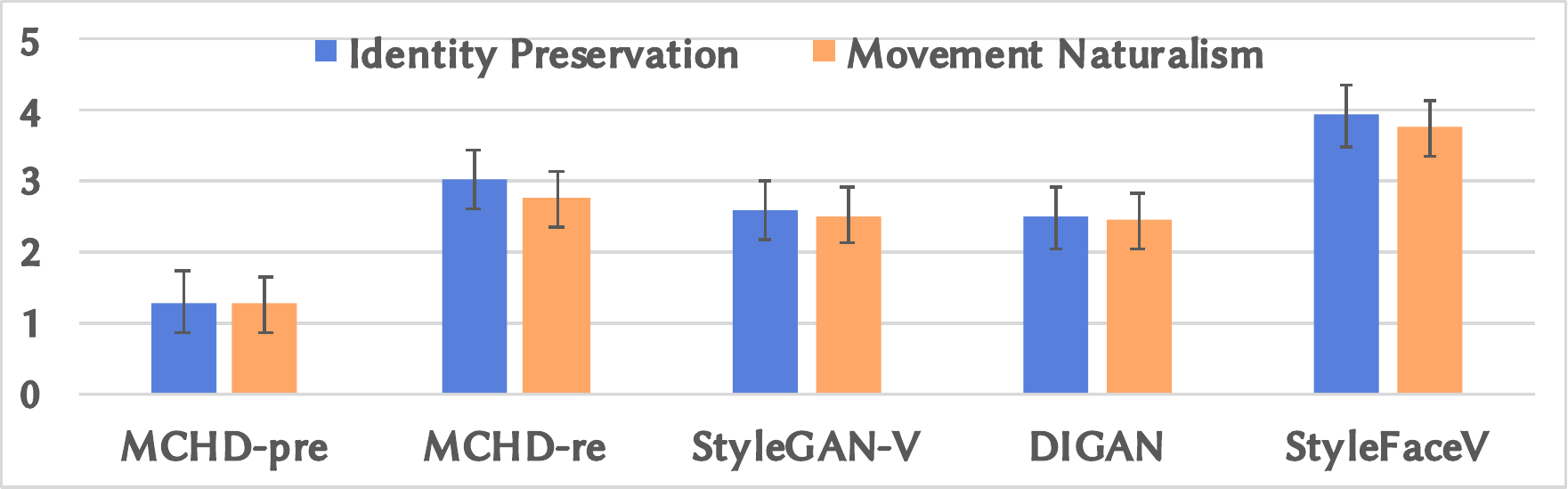}
\vspace{-5pt}
\caption{\textbf{User Study.} We asked users to give 1 - 5 scores (the higher is better) for videos generated by different methods in terms of movement naturalism and identity preservation. Our method achieves the highest score.}
\vspace{-15pt}
\label{fig:user_study}
\end{figure}

\subsection{Quantitative Comparison}

As shown in Table~\ref{table:fid_fvd}, our proposed StyleFaceV achieves the FID-RAVDESS score comparable to that of DIGAN. Since our method adopts a joint training strategy which utilizes information from both the image domain and video domain, apart from RAVDESS dataset, it is able to generate images from the FFHQ distribution. 
In other words, the synthesized frames of our method are more diverse than RAVDESS dataset. Therefore, it has a slightly higher FID-RAVDESS than DIGAN. To further evaluate if the synthesized data are truly from FFHQ distribution and RAVDESS distribution, we set up a mixture testing dataset, which is uniformly mixed with FFHQ dataset and RAVDESS dataset. We compute the FID-Mixture with this mixture dataset. As we can see, our method achieves the lowest FID-Mixture score, verifying that our method is capable of generating frames similar to both of the domains.
FVD is adopted as the metric for evaluating the quality of synthesized videos, and it is computed against two sets of videos. Compared to the state-of-the-art methods, our method has the lowest FVD score, indicating that videos synthesized by our method are more realistic.
Fig.~\ref{fig:user_study} shows the results of the user study. We compare our proposed StyleFaceV with MCHD-pre, MCHD-re, and DI-GAN. We report the preference percentage for each method. Our approach achieves the highest scores for both the naturalism of movement and identity preservation, outperforming baseline methods by a large margin.

\begin{figure}[t]
\centering
\includegraphics[width=0.8\linewidth]{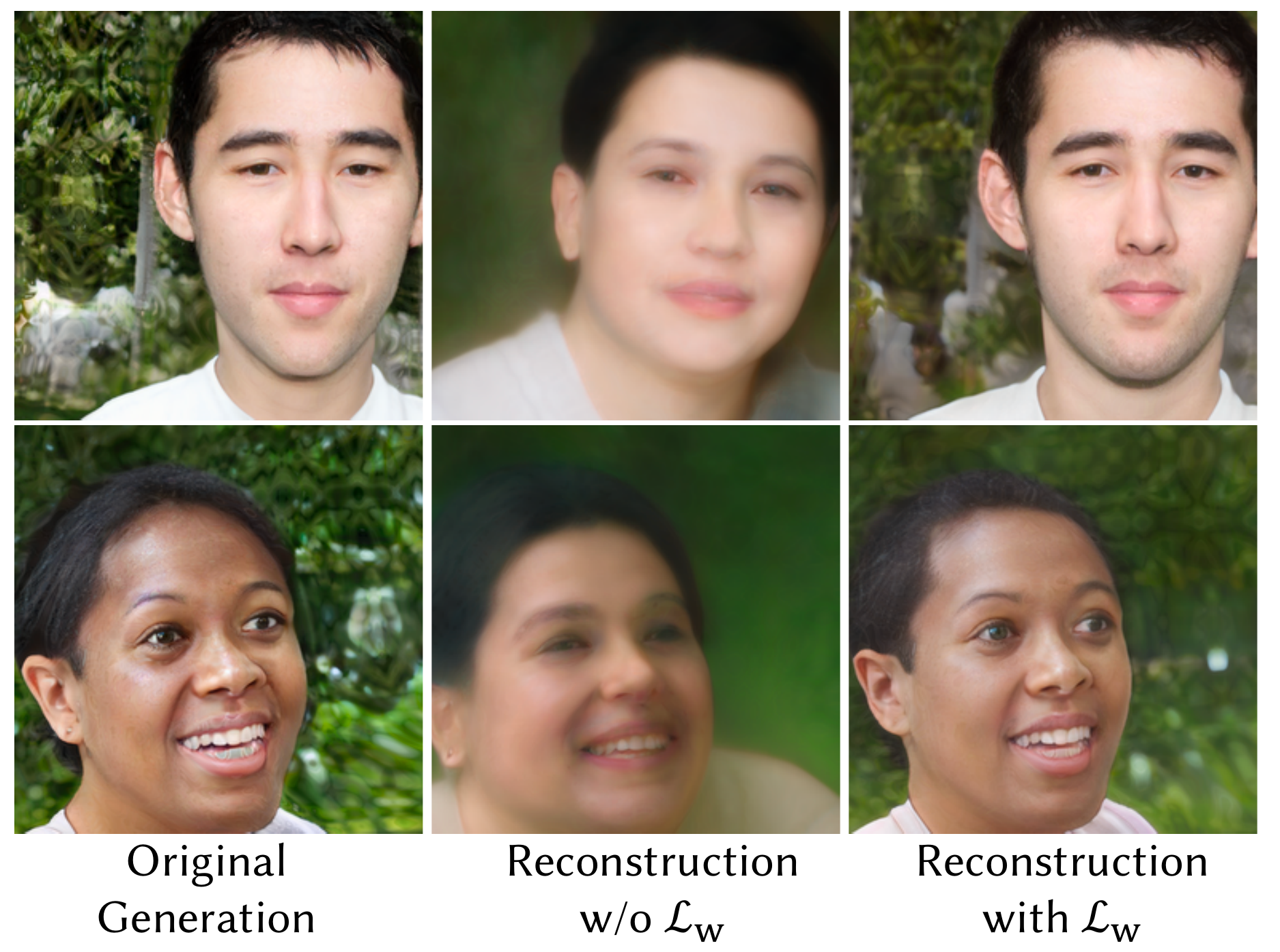}
\vspace{-10pt}
\caption{\textbf{Effectiveness of Self-Supervised Embedding Loss.} 
The images in the first column are generated by StyleGAN3. Column two and three represent reconstruction results without/with self-supervised embedding loss, respectively. The model without self-supervised embedding loss always produces severely blur results.}
\vspace{-10pt}
\label{fig:wloss}
\end{figure}

\begin{figure}[t]
\centering
\includegraphics[width=0.9\linewidth]{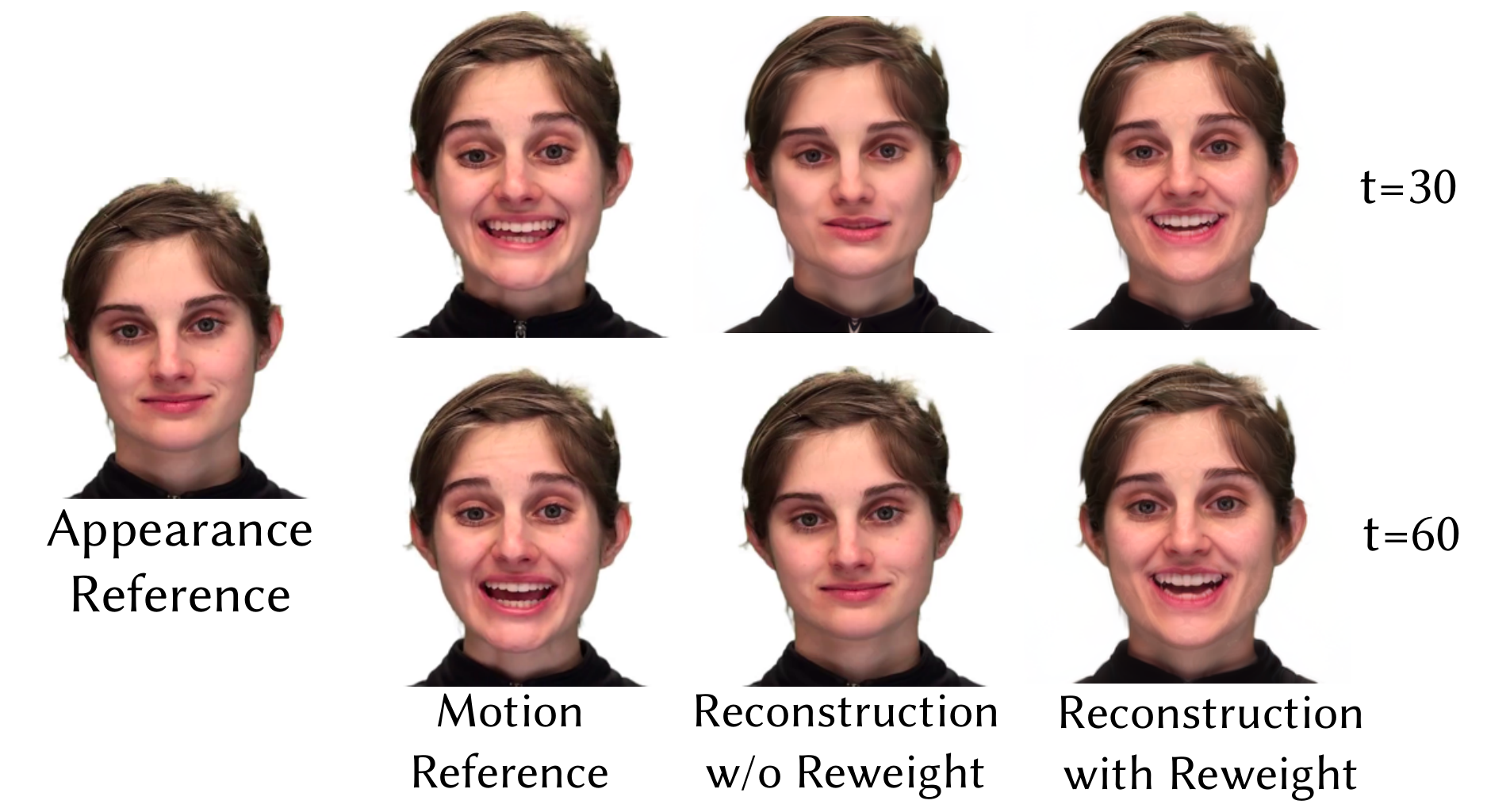}
\vspace{-10pt}
\caption{\textbf{Case without Weight Re-Balance.} 
We reconstruct images using the leftmost appearance reference and motion references.
The model without weight re-balance can not reconstruct fine details from pose representations.}
\vspace{-15pt}
\label{fig:reweight}
\end{figure}

\subsection{Ablation Study}

\begin{table}[t]
\begin{center}
\caption{\textbf{Arcface Cosine Similarity Influenced by Self-Supervised Embedding Loss.} The arcface similarity is averaged over $512$ images. The model without self-supervised embedding loss has an arcface similarity of 0.6676, while our final method achieves 0.7266 in this metric.}
\vspace{-10pt}
\label{table:arc}
\begin{tabular}{l|c}
\hline\noalign{\smallskip}
Settings & Arcface Cosine Similarity ($\uparrow$) \\
\noalign{\smallskip}
\hline
\noalign{\smallskip}
Without $\mathcal{L}_\text{w}$ & 0.6676 \\
With $\mathcal{L}_\text{w}$  & \textbf{0.7266} \\
\hline
\end{tabular}
\end{center}
\vspace{-10pt}
\end{table}
\setlength{\tabcolsep}{1.4pt}

\begin{table}[t]
\begin{center}
\caption{\textbf{Quantitative Ablations on FID and FVD Score.} We compute the FID-RAVDESS between the generated frames and real frames from RAVDESS. FID-Mixture is calculated against a mixture dataset. FVD is computed between synthesized videos and videos in RAVDESS dataset.}
\vspace{-10pt}
\label{table:abl_fid_fvd}
\scalebox{0.85}{\begin{tabular}{l|c|c|c}
\hline\noalign{\smallskip}
Methods & FID-RAVDESS ($\downarrow$) & FID-Mixture ($\downarrow$) & FVD ($\downarrow$) \\
\noalign{\smallskip}
\hline
\noalign{\smallskip}
Without Reweight & 119.90 & \textbf{21.09} & 722.84 \\
Without $\mathcal{L}_\text{w}$ & 102.97 & 67.91 & 461.17 \\
StyleFaceV & \textbf{15.42} & 25.31 & \textbf{118.72}
\\
\hline
\end{tabular}}
\end{center}
\vspace{-15pt}
\end{table}
\setlength{\tabcolsep}{1.4pt}

\noindent\textbf{Self-Supervised Embedding Loss.}
Unaligned images generated by StyleGAN3-R make models hardly converge.
Therefore, the model trained without the self-supervised embedding loss generates blurry results.
To help the model converge, we add the self-supervised embedding loss on $\mathbf{w}^*$.
Since the loss is computed directly on the latent codes, the gradients are better preserved and propagated to $F_\text{fuse}$ and $E_\text{app}$ compared to the training with only a reconstruction loss at image level.
As shown in Fig.~\ref{fig:wloss}, without the help of self-supervised embedding loss, the model is not able to recover more details of the original sampled image, thus produces severely blur results with both bad FID (FID-RAVDESS=$102.97$, FID-Mixture=$67.91$) and FVD ($461.17$).
We also report arcface cosine similarity between the reconstructed images and the original images. The arcface similarity is averaged over $512$ images. As shown in Table~\ref{table:arc} , the model without self-supervised embedding loss has an arcface similarity of 0.6676, while our final method achieves 0.7266 in this metric, further indicating the effectiveness of the self-supervised embedding loss. 

\noindent\textbf{Weight Re-Balance.}
\label{sec:ablation_rb}
At the beginning of the training of our decomposition and re-composition pipeline, the biggest challenge is the reconstruction of unaligned images with various poses. After that, the model is supposed to be capable of tracing finer details from pose representations, especially the actions of the mouth. However, the change of mouth actions is only present in video data.
To make networks focus on it, we need to re-balance the weight and make all $\lambda$s for training with sampled images significantly smaller ($\times \frac{1}{10}$). 
Fig.~\ref{fig:reweight} shows that the model with weight re-balance can successfully trace fine details from pose representations. As shown in Table.~\ref{table:abl_fid_fvd}, for the setting without Weight Re-Balance, it focus on FFHQ reconstruction (FID-RAVDESS=$119.90$) but performs poorly on RAVDESS reconstruction (FID-Mixture=$21.09$) and motion sampling (FVD=$722.84$).

\subsection{Implementation Details}

\noindent\textbf{Network structure.} $F_{app}$, $F_{pure}$, and $F_{fuse}$ are formed by res-blocks. For 2D Discriminator, we just use NLayerDiscriminator~\cite{isola2017image}. 3D Discriminator is similar but uses conv3d instead.

\noindent\textbf{Loss weights.} Initially, $\lambda_1 = 10$, $\lambda_{per} = 100$, $\lambda_{gan} = 0$, $\lambda_{pose} = 10$, $\lambda_w = 100$ for both image data and video data. After the disentanglement pipeline roughly converges and is able to reconstruct identity, we do weight rebalance to make the pose extractor mainly focus on fine motion capture. Then for image data, $\lambda_1 = 1$, $\lambda_{per} = 10$, $\lambda_{gan} = 0.01$, $\lambda_{pose} = 0.1$, $\lambda_w = 1$. For video data, $\lambda_1 = 10$, $\lambda_{per} = 100$, $\lambda_{gan} = 0.1$, $\lambda_{pose} = 1$.

\begin{figure}[t]
\centering
\includegraphics[width=0.98\linewidth]{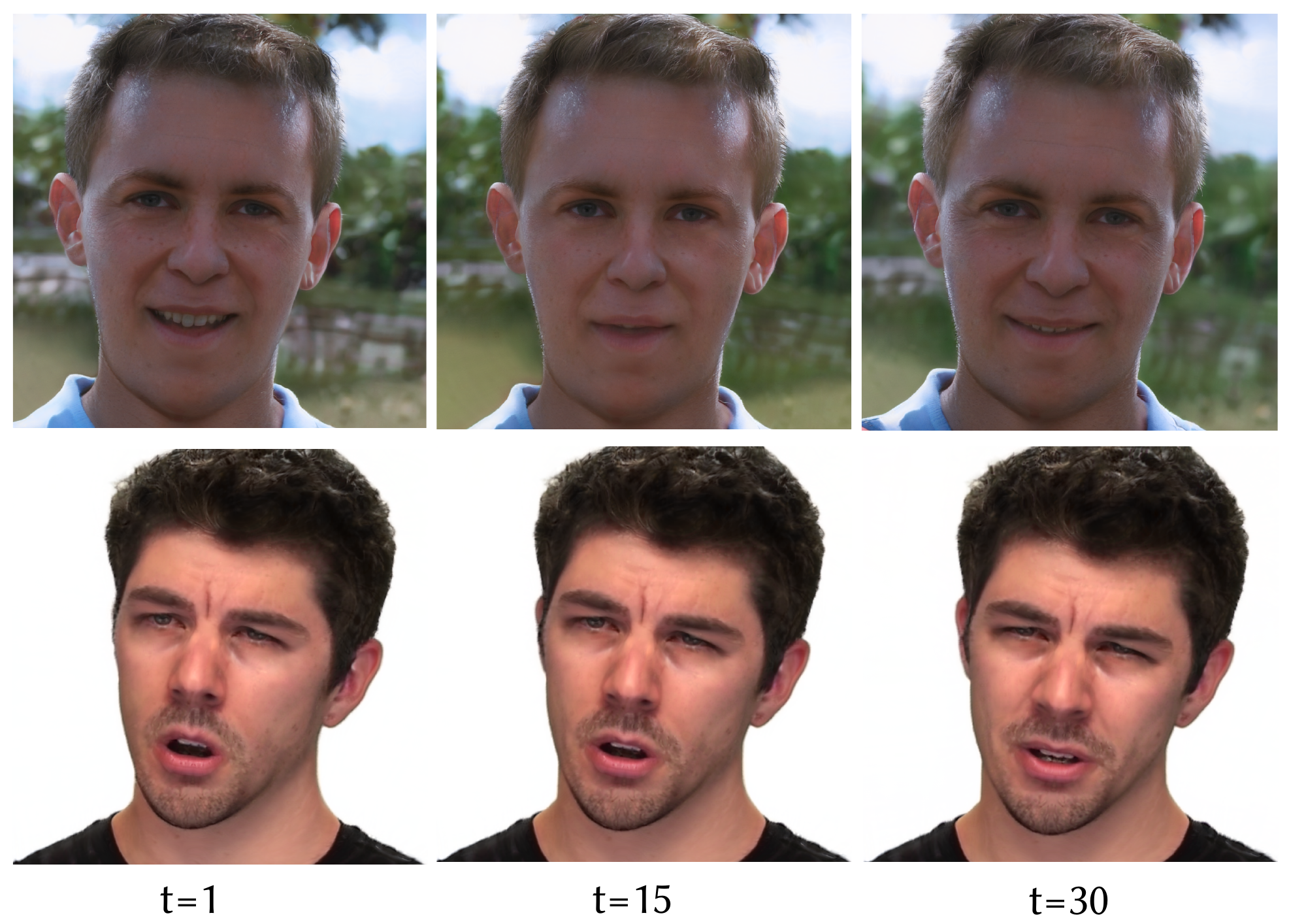}
\vspace{-10pt}
\caption{\textbf{Results of StyleFaceV in High Resolution.} 
StyleFaceV is able to generate $1024\times1024$ videos.}
\vspace{-10pt}
\label{fig:vishd}
\end{figure}

\subsection{Results in High Resolution}
We also demonstrate the capability of StyleFaceV to generate $1024\times1024$ videos in Fig.~\ref{fig:vishd}. It should be noted that the models are trained without accessing any high-resolution training videos. 

\subsection{Limitation}
Our main limitation is that the generated videos whose identity from the video domain have more vivid motions than those from the image domain. 
The reason lies in that we have direct temporal training objectives on the video domain while there is no temporal constraint for the image domain. This is a common issue for all baselines and our proposed method has already demonstrated plausible results.

\section{Conclusions}

In this work, we propose a novel framework named StyleFaceV to synthesize videos by leveraging pre-trained image generator StyleGAN3.
At the core of our framework, we decompose the inputs to the StyleGAN3 into appearance and pose information. In this way, we can generate videos from noise by sampling a sequence of pose representations while keeping the appearance information unchanged.
We also propose a joint-training strategy to make use of information from both the image domain and video domain. 
With the joint training strategy, our method is able to generate high-quality videos without large-scale training video datasets.
Extensive experiments demonstrate that our proposed StyleFaceV can generate more visually appealing videos than state-of-the-art methods.

\bibliography{aaai23}

\end{document}